\documentclass{esannV2}
\usepackage{graphicx}
\usepackage{amssymb,amsmath,array}
\usepackage{multirow}
\usepackage{booktabs}
\usepackage{makecell}
\usepackage{url}
\usepackage{subcaption}

\begin{document}
\title{FedHENet: A Frugal Federated Learning Framework for Heterogeneous Environments}

\author{Alejandro Dopico-Castro, Oscar Fontenla-Romero, Bertha Guijarro-Berdi\~nas, \\ Amparo Alonso-Betanzos and Iv\'an P\'erez Dig\'on 
\thanks{
Accepted for publication at the 34th European Symposium on Artificial Neural Networks, Computational Intelligence and Machine Learning (ESANN 2026).
\\
Work funded by Project PID2023-147404OB-I00 (MICIU/AEI/10.13039/501100011033; ERDF/EU; ESF+/EU), Horizon Europe (GA 101070381), and the Ministry for Digital Transformation and Civil Service and Next-GenerationEU/PRTR (TSI-100925-2023-1). CITIC, as a member of the CIGUS Network, receives subsidies from the ``Xunta de Galicia" and from the ERDF Operational Programme Galicia 2021-2027 (Grant ED431G 2023/01).}
%
% DO NOT MODIFY THE FOLLOWING '\vspace' ARGUMENT
\vspace{.3cm}\\
Universidade da Coru\~na, CITIC, Facultade de Inform\'atica \\ Campus de Elvi\~na s/n, A Coru\~na, Spain
}

%***********************************************************************
% END OF AUTHORS INFORMATION AREA
%***********************************************************************

\maketitle

\begin{abstract}
Federated Learning (FL) enables collaborative training without centralizing data, essential for privacy compliance in real-world scenarios involving sensitive visual information. Most FL approaches rely on expensive, iterative deep network optimization, which still risks privacy via shared gradients. In this work, we propose FedHENet, extending the FedHEONN framework to image classification. By using a fixed, pre-trained feature extractor and learning only a single output layer, we avoid costly local fine-tuning. This layer is learned by analytically aggregating client knowledge in a single round of communication using homomorphic encryption (HE). Experiments show that FedHENet achieves competitive accuracy compared to iterative FL baselines while demonstrating superior stability performance and up to 70\% better energy efficiency. Crucially, our method is hyperparameter-free, removing the carbon footprint associated with hyperparameter tuning in standard FL. Code available in \url{https://github.com/AlejandroDopico2/FedHENet/}
\end{abstract}

\section{Introduction}

Federated learning (FL) has emerged as a crucial paradigm for collaborative model training without sharing sensitive data. However, foundational algorithms like FedAvg \cite{mcmahan2017communication} or FedProx \cite{li2020federated} rely on the iterative optimization of deep neural networks. This leads to three major drawbacks: (1) poor convergence and client drift in heterogeneous scenarios; (2) high communication and computation costs over hundreds of rounds; and (3) vulnerability to privacy attacks via shared gradients.

To address these limitations, we propose FedHENet, a framework designed for ``Frugal FL" in computer vision. While FedHEONN \cite{fontenla2023federated} validated single-round, homomorphically encrypted (HE) learning for tabular data, extending this to high-dimensional image domains is non-trivial. FedHENet solves this by decoupling feature extraction (via a pre-trained, frozen backbone) from the learning process. Instead of costly local fine-tuning, clients compute a lightweight, analytically solvable output layer (ROLANN). This ensures mathematically exact global aggregation in a single round, eliminating the need for hyperparameter tuning and drastically reducing the environmental footprint.

Our main contributions are:

\begin{itemize}
    \item \textbf{Frugal adaptation to image data:} A hybrid architecture combining frozen feature extractors with an analytical layer whose transmitted statistics are homomorphically encrypted, enabling secure single-round convergence on image tasks without iterative gradient descent.
    \item \textbf{Superior stability and efficiency:} We demonstrate robustness against extreme non-IID data (achieving high accuracy where iterative baselines fail) with up to 70\% energy savings.
    \item \textbf{Scalable deployment:} We integrate MQTT for fault-tolerant communication and provide an open-source library to facilitate encrypted Frugal FL deployment.
\end{itemize}

\section{Method}

FedHENet combines a fixed, pre-trained (on ImageNet) feature extractor with a Regularized One-Layer Neural Network (ROLANN) classifier \cite{fontenla2021rolann}. All clients share the same frozen feature extractor to produce compact embeddings $X_{k} \in \mathbb{R}^{m_{k} \times n}$ from their local $m_k$ samples. By not performing any fine-tuning, we avoid high computational and energy costs.

On top of these features, each client trains a ROLANN layer. Unlike traditional FL, this layer computes its optimal weights in a closed form, not via iterative gradient descent. This is achieved by minimizing the MSE measured \emph{before} the activation function on the desired pre-activation outputs $d_k$, which allows to transform the regularized least-squares problem into a system of linear equations that can be solved analytically, enabling single-round aggregation.

Each client $k$ calculates the required components. The activation derivative $f'$ is calculated on the desired pre-activaction outputs $\bar{d}_k$. Client $k$ computes $f'$, forms a diagonal matrix $F_k = \text{diag}(f'_k)$, and performs:

\begin{equation}
    [U_k, S_k, \sim] = \text{SVD}(X_k F_k), \quad M_k = X_k(f'_k \odot f'_k \odot \bar{d}_k),
\end{equation}

The only component containing sensitive statistical dependencies from the training data is the matrix $M_k$, defined as a weighted correlation between features and pre-activation terms. To protect $M_k$, each client encrypts it using the CKKS homomorphic encryption \cite{cheon2017homomorphic}, producing $[[M_k]]$. CKKS is used because it enables arithmetic operations on approximate (real) numbers, essential for this analytical aggregation.  %In contrast, $U_k$ and $S_k$ are safe to transmit in plaintext. % because they do not permit reconstruction of the raw data.

Clients send ($U_k, S_k, [[M_k]]$) to the coordinator using MQTT, which provides lightweight, asynchronous, and fault-tolerant communication. $U_k$ and $S_k$ are transmitted in plaintext and concatenated at the server to compute the global singular values ($U, S$). The matrices $M_k$ are sent encrypted and their statistics are aggregated homomorphically:

\begin{equation}
    [U, S, \sim] = \text{SVD}([U_1, S_1 \|\, U_2, S_2 \|\, \dots \|\, U_K, S_K), \quad [[M]] = \sum_{k=1}^{K} [[M_k]]
\end{equation}

The coordinator computes the global ROLANN weights $W$ from $U$, $S$ and $[[M]]$ in a single analytical step and publishes $W$ to all clients using MQTT. This avoids redundant local computation, since all clients would obtain the same weights.

\begin{equation}
    W = U \cdot (S \cdot S + \lambda I)^{-1}U^T[[M]]
\end{equation}

This closed-form expression yields the globally optimal classifier without iterative refinement and ensures exact aggregation across heterogeneous clients.

\section{Experimental Results and Discussion}

\paragraph{Experimental Setup}
We implemented the framework\footnote{\url{https://github.com/AlejandroDopico2/FedHENet/}} in PyTorch and executed on a workstation (Intel Core i7-10700KF, NVIDIA RTX 3080). Communication relied on MQTT (Eclipse Mosquitto) and CKKS Homomorphic encryption. Experiments used CIFAR-10 and CIFAR-100 datasets, containing 10 and 100 classes of $32\times32$ color images following the standard train and test splits, respectively. Data heterogeneity was simulated using a Dirichlet distribution ($\alpha$) where lower $\alpha$ signifies non-IID data. Additionally, a ``single-class" scenario (where each client only gets data for one class) was used to represent extreme heterogeneity. We compare FedHENet against FedAvg and FedProx using the same frozen ResNet-18 backbone. Baselines were trained for 10 global rounds on CIFAR-10 and 50 global rounds on CIFAR-100 with 1 local epoch per round and full client participation. These round counts were empirically determined to ensure the iterative algorithms reached a stable, near-optimal convergence for a fair performance comparison. Metrics include test accuracy, energy consumption (measured via CodeCarbon\footnote{\url{https://github.com/mlco2/codecarbon}}), training time, and cumulative communication volume (the total data transmitted by all participants across the entire process).

\paragraph{Accuracy and Robustness Analysis}

Table \ref{tab:fedhenet-accuracy} highlights the test results. While iterative methods (FedAvg and FedProx) perform well on CIFAR-10 in near-IID settings ($\alpha = 1.0$) with few client ($N=10$), their performance deteriorates as data heterogeneity increases. In the extreme ``single class" scenario ($N=10$), baselines drop to $33-40\%$ accuracy due to severe client drift. FedHENet, relying on exact analytical aggregation, remains immune to drift, maintaining $\sim 83.68\%$ accuracy. In the large-scale experiment with $N=500$ clients, our approach outperforms FedAvg by over $8\%$ in the non-IID setting ($\alpha=0.1$).

On the complex CIFAR-100 dataset, FedHENet achieves competitive results ($\sim 56\%$) against baselines that needed 50 rounds to achieve stable convergence. In highly heterogeneous settings, FedHENet matches or beats the baselines, which suffer from divergence despite the extended training.

\begin{table}[t]
\centering
\small
\begin{tabular}{lllccc}
\toprule
Dataset & $N$ & $\alpha$ & FedAvg & FedProx & \textbf{FedHENet} \\
\midrule
\multirow{10}{*}{CIFAR-10}  & \multirow{3}{*}{10}   & 1.0 & \textbf{85.34}  & 85.28   & 83.69    \\
                           &                           & 0.1 & 81.25  & 78.52   & \textbf{83.68}    \\
                           &                           & single       & 33.26  & 40.75   & \textbf{83.65}    \\
\cmidrule(lr){2-6} 
                           & \multirow{3}{*}{100} & 1.0 & 79.82  & 79.84   & \textbf{83.64}    \\
                           &                           & 0.1 & 76.44  & 76.89   & \textbf{83.66}    \\
                           &                           & single       & 48.12  & 51.19   & \textbf{83.63}    \\
\cmidrule(lr){2-6} 
                           & \multirow{3}{*}{500}     & 1.0 &  73.78      &   73.78      &     \textbf{83.79}     \\
                           &                           & 0.1 &    75.18    &    75.18     &   \textbf{83.59}       \\
                           &                           & single      &     73.26   &   73.28      &       \textbf{83.70}   \\
\midrule 
\multirow{3}{*}{CIFAR-100} & \multirow{3}{*}{100} & 1.0 &   \textbf{59.11}     & 59.08         &     56.63     \\
                           &                           & 0.1 &    \textbf{58.2}    &  58.15        &      56.13    \\
                           &                           & single      &    52.61    &    52.6     &    \textbf{56.91}      \\
\bottomrule
\end{tabular}
\caption{Top-1 test accuracy (\%) comparison between FedAvg, FedProx, and FedHENet, under a varying number of clients ($N$) and data heterogeneity ($\alpha$).}
\label{tab:fedhenet-accuracy}
\end{table}

\paragraph{Frugal Efficiency and Sustainability}

Table \ref{tab:fedhenet-green-metrics} confirms massive resource reductions. For CIFAR-10 ($N=10$), FedHENet reduces energy by $\sim 70\%$ ($11.5$ Wh vs $36.5$ Wh). Although FedHENet’s single payload is larger than a standard gradient update, the overall byte count is significantly lower than the baselines. This gap widens for CIFAR-100 where baselines consumed $\sim240$ Wh (due to the 50-round requirement), while FedHENet required only 118.9 Wh, halving the cost.

\begin{table}[t]
\centering
\small
\begin{tabular}{llcccc}
\toprule
Dataset & $N$ & \makecell{Metric} & FedAvg & FedProx & \textbf{FedHENet} \\
\midrule
\multirow{10}{*}{CIFAR-10}
& \multirow{3}{*}{10} & Time (min.) & 8.55 & 10.11 & \textbf{3.07} \\
& & Energy (Wh) & 30.2 & 36.5 & \textbf{11.5} \\
& & Bytes (MB) & 358.03 & 358.04 & \textbf{331.33} \\
\cmidrule{2-6}
& \multirow{3}{*}{100} & Time (min.) & 12.15 & 14.37 & \textbf{3.77} \\
& & Energy (Wh) & 40.6 & 47.0 & \textbf{11.2} \\
& & Bytes (MB) & 3369.84 & 3369.85 & \textbf{2855.51} \\
\cmidrule{2-6}
& \multirow{3}{*}{500} & Time (min.) & 32.80 & 32.72 & \textbf{12.32} \\
& & Energy (Wh) & 96.85 & 95.07 & \textbf{36.01} \\
& & Bytes (MB) & 16755.52 & 16755.50 & \textbf{9694.33} \\\midrule
\multirow{3}{*}{CIFAR-100} & \multirow{3}{*}{100} & Time (min.) & 77.71 & 81.98 & \textbf{40.32} \\
& & Energy (Wh) & 236.01 &  248.93 & \textbf{118.9} \\
& & Bytes (MB) & 42121.21 & 42121.21 & \textbf{29615.90} \\
\bottomrule
\end{tabular}
\caption{Energy, communication and time efficiency comparison (`Green metrics') of FedAvg, FedProx, and FedHENet for CIFAR datasets.}
\label{tab:fedhenet-green-metrics}
\end{table}

A crucial, often overlooked aspect of sustainability is the cost of hyperparameter tuning. Iterative methods require energy-intensive grid searches (LR, decay schedules, and local epochs) to prevent divergence. This search phase multiplies the true energy cost of deployment. FedHENet is hyperparameter-free, eliminating this search phase and drastically reducing the carbon footprint of the entire development lifecycle, not just the final training run.

\paragraph{Accuracy-Energy Trade-off}

\begin{figure}[!h]
    \centering

    \begin{subfigure}[t]{0.48\linewidth}
        \centering
        \includegraphics[width=\linewidth]{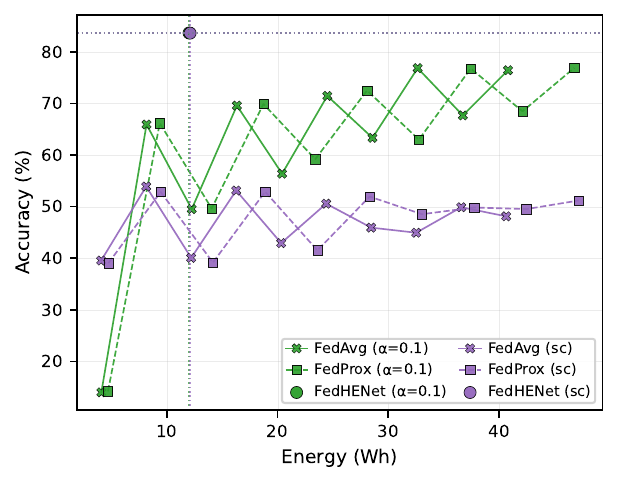}
    \end{subfigure}
    \hfill
    \begin{subfigure}[t]{0.48\linewidth}
        \centering
        \includegraphics[width=\linewidth]{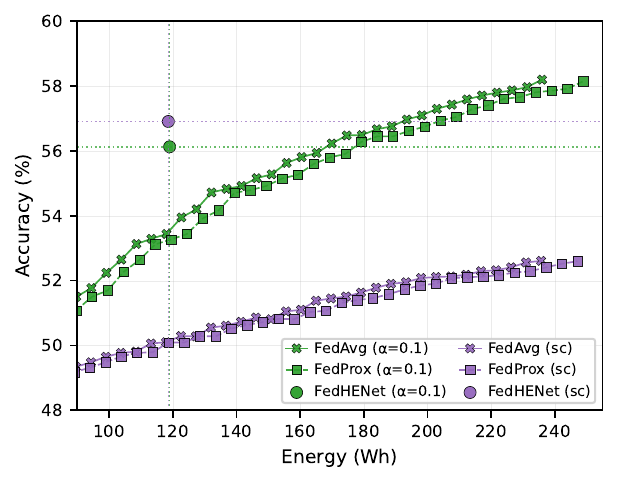}
    \end{subfigure}

    \caption{Accuracy vs. energy consumption per training round for CIFAR-10 (left) and CIFAR-100 (right) for $\alpha = 0.1$ and single-class (sc) scenarios with 100 clients. FedHENet achieves peak accuracy in a single round, while baselines require multiple rounds. In CIFAR-10, the two circles overlap. For clarity, in CIFAR-100 only the last 30 baseline rounds are shown.}
    \label{fig:main-figure}
\end{figure}

The plot in Figure \ref{fig:main-figure}, illustrating Accuracy vs. Energy consumption per round, confirms FedHENet’s frugality. The iterative baselines are forced to continue consuming energy over multiple rounds to asymptotically approach FedHENet's single-round accuracy. For example, in CIFAR-10, FedHENet achieves $\sim 83.6\%$ accuracy using only $\sim 11.5$ Wh of energy, while iterative methods consume over three times the energy ($\sim 40$ Wh) to converge. Furthermore, their performance in the non-IID setting ($\alpha=0.1$) is highly unstable, oscillating wildly between $45\%$ and $77\%$ accuracy across rounds, demonstrating low robustness despite high energy expenditure. This frugality is magnified on CIFAR-100, where the single-round execution provided an energy reduction of nearly half compared to baselines that required extensive, multi-round training.

% \begin{figure}[!h]
%     \centering

%     \begin{subfigure}{0.48\linewidth}
%         \centering
%         \includegraphics[scale=0.6]{accuracy_vs_energy_cifar10_dir_a0.1_nc100_sc_a1.0_nc100.eps}
%     \end{subfigure}
%     \hfill
%     \begin{subfigure}{0.48\linewidth}
%         \centering
%         \includegraphics[scale=0.6]{accuracy_vs_energy_cifar100_dir_a0.1_nc100_sc_a1.0_nc100.eps}
%     \end{subfigure}
%     \caption{Accuracy vs. energy consumption per training round for CIFAR-10 (left) and CIFAR-100 (right) for the $\alpha$ = 0.1 and single class (sc) scenarios with 100 clients. FedHENet achieves peak accuracy in a single round, while baselines require multiple rounds. In CIFAR-10, the two circles overlap. For clarity, in CIFAR-100 only the last 30 baseline rounds are shown.}
%     \label{fig:main-figure}
% \end{figure}

\paragraph{Homomorphic Encryption Overhead}

Encryption increases the payload size ($M_k$ matrix) by approximately $2.25 \times$, increasing transmitted bytes (Table \ref{tab:he-vs-nohe}). However, the computational time overhead is negligible ($<5\%$). This confirms that FedHENet provides robust privacy via without compromising its frugal nature. While this overhead remains negligible for a moderate number of clients, it could become more noticeable in massive-scale federations, where encryption and communication costs scale linearly with the client count.

\begin{table}[t]
\centering
\begin{tabular}{lcccccc}
\toprule
FedHENet & \makecell{Acc.\\(\%)} & \makecell{Bytes\\(MB)} & \makecell{CT size\\inflation} & \makecell{Energy\\(Wh)} & \makecell{Time\\(min.)} & \makecell{Overhead\\(ms/client)} \\
\midrule
\quad w/o HE & 83.69 & 147.6 & 1.0$\times$ & 10.60 & 2.92 & -- \\
\quad  w/ HE & 83.69 & 331.33 & 2.25$\times$ & 10.93 & 3.06 & 840 \\
\bottomrule
\end{tabular}
\caption{Impact of applying Homomorphic Encryption (HE) to FedHENet on CIFAR-10 ($\alpha$=1.0, N=10).}
\label{tab:he-vs-nohe}
\end{table}

\section{Conclusions}

We introduced FedHENet, a framework that addresses resource constraints and heterogeneity via a computationally frugal, single-round design. By decoupling feature extraction from a closed-form learning layer, it achieves up to $70\%$ energy savings compared to iterative baselines. Our results demonstrate that, unlike gradient-based methods, FedHENet avoids client drift in extreme non-IID settings. Furthermore, by being hyperparameter-free, it offers a sustainable path for edge AI, eliminating the energy-intensive tuning phase required by standard approaches. The production-ready core implementation of this framework, which is designed for seamless deployment in a client network, is available as part of a general-purpose federated learning library repository.

For future work, we plan to validate FedHENet on heterogeneous edge hardware (e.g., Raspberry Pi clusters) to precisely measure real-world performance metrics. We will also explore the use of other foundational backbones, such as Vision Transformers.

\begin{footnotesize}

\bibliographystyle{unsrt}
\bibliography{bib}

\end{footnotesize}

\end{document}